\documentclass[10pt,twocolumn,letterpaper]{article}

\usepackage{cvpr}              

\usepackage[dvipsnames]{xcolor}

\definecolor{cvprblue}{rgb}{0.21,0.49,0.74}
\usepackage[pagebackref,breaklinks,colorlinks,citecolor=cvprblue]{hyperref}
\usepackage{multirow}
\usepackage{colortbl}
\usepackage{algorithm}
\usepackage{algorithmic}
\usepackage{xcolor}

\def\paperID{11526} 
\def\confName{CVPR}
\def\confYear{2024}

\title{AutoAugment Input Transformation for Highly Transferable Targeted Attacks}

\author{
Haobo Lu\thanks{The first two authors contributed equally. Correspondence to Kun He.}, Xin Liu\footnote[1]{}, Kun He\\ 
School of Computer Science and Technology, \\
Huazhong University of Science and Technology, Wuhan, China\\
{\tt\small \{haobo,liuxin\_jhl, brooklet60\} @hust.edu.cn
}}

\begin{document}
\maketitle
\begin{abstract}
Deep Neural Networks (DNNs) are widely acknowledged to be susceptible to adversarial examples, wherein imperceptible perturbations are added to clean examples through diverse input transformation attacks. 
However, these methods originally designed for non-targeted attacks exhibit low success rates in targeted attacks. 
Recent targeted adversarial attacks mainly pay attention to gradient optimization, attempting to find the suitable perturbation direction.
However, few of them are dedicated to input transformation.
In this work, we observe a positive correlation between the logit/probability  of the target class and diverse input transformation methods in targeted attacks.
To this end, we propose a novel targeted adversarial attack called AutoAugment Input Transformation (AAIT).
Instead of relying on hand-made strategies, AAIT searches for the optimal transformation policy from a transformation space comprising various operations.
Then, AAIT crafts adversarial examples using the found optimal transformation policy to boost the adversarial transferability in targeted attacks.
Extensive experiments conducted on CIFAR-10 and ImageNet-Compatible datasets demonstrate that the proposed AAIT surpasses other transfer-based targeted attacks significantly.
\end{abstract}    
\section{Introduction}

\label{sec:intro}
\begin{figure*}[t]
  \centering
   \includegraphics[width=0.9\linewidth]{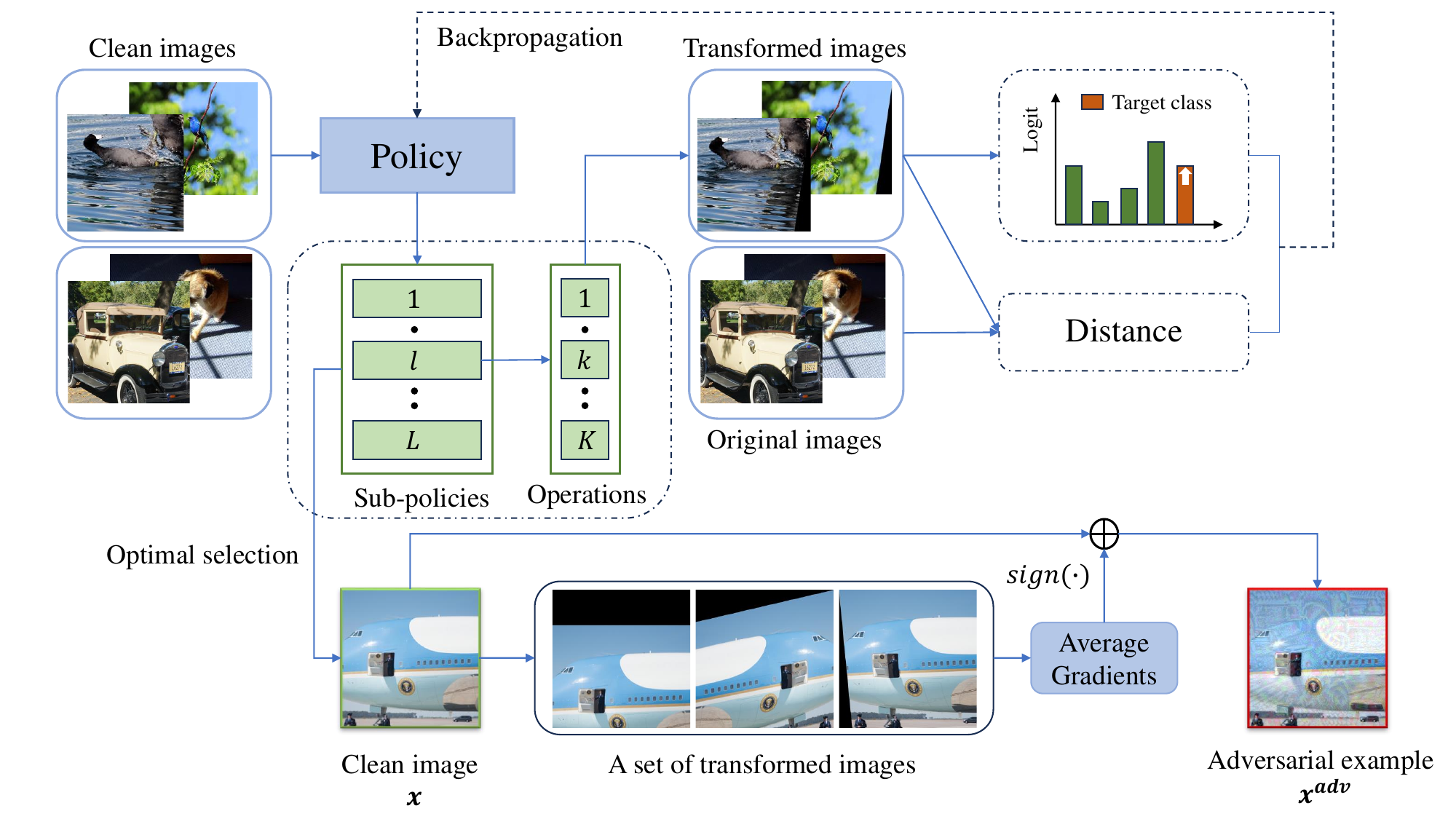}
   \caption{The overview of the proposed AAIT method. We search for an optimal policy according to the logit and distance between the original and transformed images. We then craft adversarial examples using the optimal policy to obtain the average gradients from a set of transformed images.}
   \label{fig:overview}
\end{figure*}

Deep Neural Networks (DNNs) have outperformed humans on enormous  tasks~\cite{resnet,DBLP:conf/nips/KrizhevskySH12,bert,Attention}.
However, DNNs are widely known to be vulnerable to adversarial examples by adding human-imperceptible perturbations to clean examples.
Existing works~\cite{FGSM} have demonstrated that adversarial examples generated on the white-box model can deceive other black-box models, called adversarial transferability.
It is necessary to explore effective attack methods, not only for attacking target models, but also be useful for the model robustness evaluation.

Adversarial attacks can be divided into two typical categories, \ie, white-box attacks~\cite{FGSM,IFGSM} and black-box attacks~\cite{MIM}.
Existing works focus more on black-box attacks since the actual scenario is more consistent with the black-box settings.
One of the most effective and efficient black-box attacks is transfer-based attacks~\cite{MIM,NI,VT}.
Recently, transfer-based methods have achieved high attack success rates in non-targeted attacks.
However, they still exhibit low attack success rates in targeted attacks, which makes the victim model classify the adversarial example to a specific target class.
For instance, 
we observe that input transformation attacks that are originally designed for non-target attacks, such as the Scale-Invariant method (SIM)~\cite{NI} and Admix~\cite{Admix}, perform unsatisfactorily while directly adapting to targeted attacks.

To improve the success rates of targeted adversarial attacks, some works have designed better loss functions, such as Logit Loss~\cite{Logit_loss} and Po+Trip Loss~\cite{Po+Trip_loss}, to overcome the vanishing gradient caused by Cross-Entropy (CE) loss during large iterations.
Other methods~\cite{FDA,FDA+xnet,TTP} try to align the feature of the generated adversarial example with the feature distribution of the target class.
Nonetheless, few input transformation attacks have been explored in targeted adversarial attacks.

In this work, we investigate the targeted input transformation method to further improve the transferability of adversarial examples in targeted attacks.
Specifically, we observe a positive correlation between the logit/probability of the target class and diverse input transformation methods in targeted attacks.
Therefore, it is wise to explore effective input transformation methods that can amplify the logit/probability of the target class.
Note that existing methods (e.g., DIM~\cite{DI}, SIM~\cite{NI}) are all hand-made, which may limit the potential capability of the input transformation methods.
To address this issue, we introduce automatic input transformation 
for targeted attacks. 

Specifically, we propose a novel targeted adversarial attack called AutoAugment Input Transformation (AAIT).
The overview of AAIT is illustrated in Figure~\ref{fig:overview}.
Specifically, AAIT first picks clean examples and divides them into two parts.
Then, it searches for an optimal input transformation policy that maximizes the logit of the transformed examples while minimizing the distance of distributions between the original and transformed examples. 
Finally, we obtain the average gradient on a set of transformed images by AAIT to craft more transferable adversarial examples. 
In summary, our main contributions are as follows:
\begin{itemize}
    \item We observe a clear positive correlation between the logit of the target class and diverse input transformation methods in targeted attacks. Based on this observation, we propose a novel targeted adversarial attack framework called AutoAugment Input Transformation (AAIT).
    \item Instead of relying on hand-made strategies, AAIT searches for the optimal input transformation policy through a search algorithm, which maximizes the logit of the transformed examples while minimizing the distribution distance between the original and transformed examples. Then, it crafts adversarial examples with the optimal policy to transform images.
    \item Extensive experiments conducted on CIFAR-10 and ImageNet-Compatible datasets demonstrate that AAIT outperforms other transfer-based targeted attacks by a clear margin. AAIT can efficiently search for the optimal input transformation and significantly enhance the transferability of adversarial examples.
\end{itemize}
\section{Related Work}
This section reviews the definition of adversarial attacks and previous works on non-targeted and targeted transferable attacks.

\subsection{Adversarial Attacks}

Let \(\boldsymbol{x}\) and \(y\) be a clean example and the corresponding true label, respectively.
The \(J(f(\boldsymbol{x}), y)\) is defined as the loss function of the model.
The targeted adversarial attack aims to generate an adversarial example, denoted as \(\boldsymbol{x}^{adv}\), by minimizing the loss function as follows:
\begin{equation}
    \arg \min_{\boldsymbol{x}^{adv}} \boldsymbol{J}(f(\boldsymbol{x}^{adv}), y_t)
    \quad s.t. \quad \|\boldsymbol{x}-\boldsymbol{x}^{adv}\|_\infty \leq \epsilon ,
\end{equation}
where \(y_t\) is the target class and \(\epsilon\) is the magnitude of perturbation.
One of the most classic white-box adversarial attacks is Fast Gradient Sign Method (FGSM)~\cite{FGSM}, which crafts an adversarial example by utilizing the sign of the input gradient as follows:
\begin{equation}
    \boldsymbol{x}^{adv}=\boldsymbol{x}-\epsilon \cdot sign (\nabla_{\boldsymbol{x}} \boldsymbol{J}(f(\boldsymbol{x}),y_t)),
\end{equation}
where \(sign(\cdot)\) is the sign function.
Iterative Fast Gradient Sign Method (I-FGSM)~\cite{IFGSM} extends the FGSM to the iteration version, which generates adversarial examples with multiple iterations and smaller step sizes.

\subsection{Non-targeted Transferable Attacks}
Numerous works have been proposed to enhance the transferability of adversarial examples.
Among them, gradient optimization and input transformation are the most effective methods.
Gradient optimization attacks introduce new gradient calculation methods or loss functions to improve the transferability of adversarial examples.
Momentum Iterative Fast Gradient Sign Method (MI-FGSM)~\cite{MIM} and Nesterov Iterative Fast Gradient Sign Method (NI-FGSM)~\cite{NI} introduce momentum and Nesterov term to accelerate the gradient convergence.
Variance Tuning (VT)~\cite{VT} fuses the gradients generated by the neighborhood of current data to decrease variance.
Attention-guided Transfer Attack (ATA)~\cite{ATA} obtains the attention weights of features by Grad-CAM~\cite{Grad}.
Feature Importance-aware Attack (FIA)~\cite{FIA} averages feature gradients of various augmented images as attention weights.
Input transformation attack generates adversarial perturbation with various transformation patterns.
Diverse Input Method (DIM)~\cite{DI} transforms the input image with random resizing and padding.
Translation-Invariant Method (TIM)~\cite{TIM} smooths gradients through a predefined convolution kernel to overcome the overfitting.
Scale-Invariant Method (SIM)~\cite{NI} calculates the average gradient of inputs across different scales.
Admix~\cite{Admix} extends the SIM to improve the attack transferability, mixing images from other labels.
Nonetheless, the performance of these methods is unsatisfactory in the setting of targeted attack.

\subsection{Targeted Transferable Attacks}
Previous works are dedicated to learning feature distributions of the targeted class.
Feature Distribution Attack (FDA)~\cite{FDA} maximizes the probability outputted from a binary classifier, which is trained by utilizing the intermediate features of the training dataset from the white-box model.
FDA$^{(N)}$+xent~\cite{FDA+xnet} extends FDA by incorporating the CE loss and aggregating features from multi-layers.
Transferable Targeted Perturbations (TTP)~\cite{TTP} trains a generator function that can adaptively synthesize perturbations specific to a given input.

Po+Trip loss~\cite{Po+Trip_loss} first demonstrates the gradient vanishing problem caused by CE loss and introduces the Poincar\'e distance to measure the similarity.
Logit loss~\cite{Logit_loss} utilizes the logit of the targeted class as the classification loss function and enlarges the number of iterations to achieve better performance.
Object-based Diverse Input (ODI)~\cite{ODI} projects an adversarial example on 3D object surfaces to improve input diversity.
Self-Universality (SU) attack~\cite{Self-Universality} maximizes the feature similarity between global images and randomly cropped local regions.
Clean Feature Mixup (CFM)~\cite{CFM} randomly mixes stored clean features with current input features, effectively mitigating the overfitting issue.

\subsection{The AutoAugment Family}
AutoAugment~\cite{AutoAugment} is the first method introduced to design an automated search for augmentation strategies directly from a dataset. 
It creates a search space that consists of many image processing functions. 
The purpose is to search for a best combination.
To deal with the long search time of AutoAugment, Population-Based Augmentation (PBA)~\cite{PBA} generates augmentation policy schedules based on population-based training.
Fast AutoAugment~\cite{Fast} uses Bayesian optimization in the policy search phase, speeding up the search time by orders of magnitude.
Faster AutoAugment~\cite{Faster} proposes a differentiable policy search pipeline that uses backpropagation to update policy.
In this paper, we follow the framework of Faster AutoAugment to spend less search time. 
Note that the original AutoAugment is for model training to improve the generalization, but we adopt it to enhance the transferability of targeted attacks.

\section{Methodology}

This section first introduces our motivation. Then, we rethink operations used in AutoAugment. 
Finally, we present our AutoAugment Input Transformation (AAIT) method to overcome the limitation of existing targeted attacks.

\subsection{Motivation}
Admix~\cite{Admix} has achieved high attack success rates in non-targeted attacks, which adds a small portion of other class images and makes the clean example closer to the decision boundary.
For targeted attacks, we argue that 
the information of the target class is more important for enhancing the transferability of adversarial examples. 

\begin{figure*}[t]
  \centering
  \begin{subfigure}{0.45\linewidth}
    \includegraphics[width=1\linewidth]{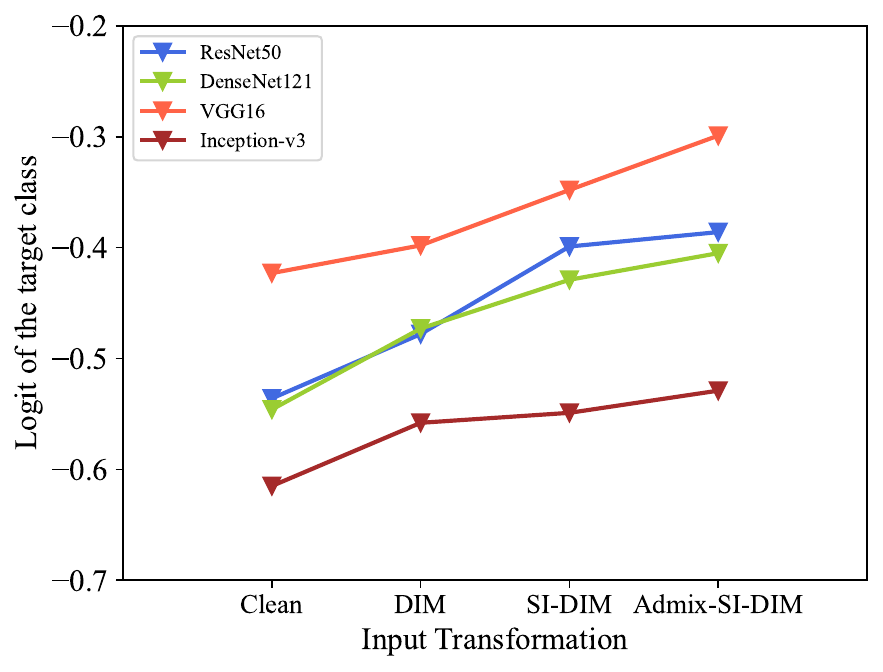}
    \caption{The logit of different input transformation attacks.}
    \label{fig:short-a}
  \end{subfigure}
  \hfill
  \begin{subfigure}{0.45\linewidth}
    \includegraphics[width=1\linewidth]{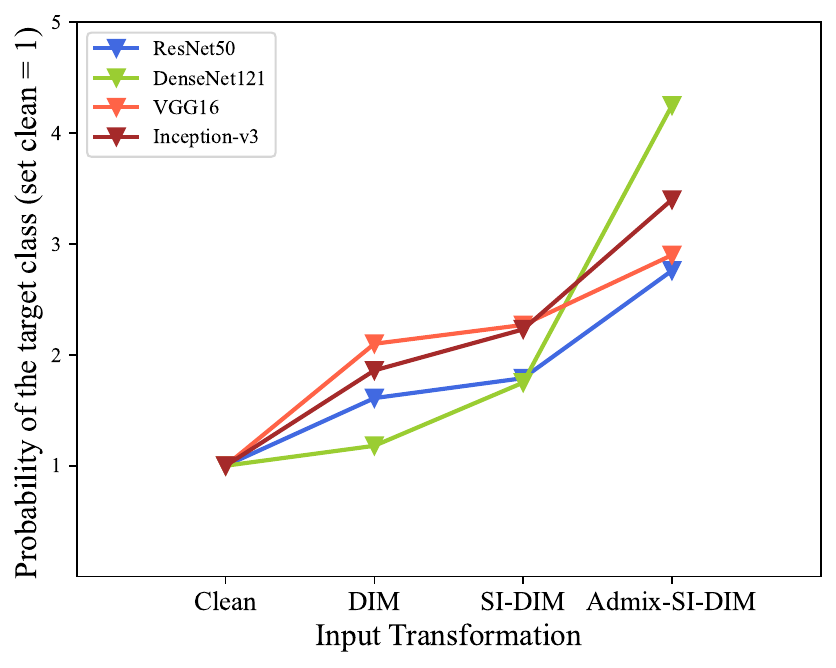}
    \caption{The probability of different input transformation attacks.}
    \label{fig:short-b}
  \end{subfigure}
  \caption{The relationship between logit/probability of the target class and different input transformation methods to transform images. 
  Note that the logit/probability and diverse input transformation methods exhibit a clear positive correlation.
  a) The combination of different input transformation methods increases the logit of the target class.
  b) The probability of transformed images using diverse input transformation methods exhibits a clear positive correlation.
  }
  \label{fig:motivation}
\end{figure*}

To verify our assumption, we conduct experiments using several typical input transformation methods to explore the relationship between the logit/probability of the target class and input transformation diversity.
Specifically, we choose DIM, SIM, and Admix and combine them as SI-DIM and Admix-SI-DIM.
As shown in Figure~\ref{fig:short-a}, we observe a clear positive correlation between the logit and input transformation diversity. 
The targeted attack success rate has significantly improved by combining different input transformations.
Furthermore, to enhance the credibility of our results, we have also provided the probability of the target class, which is obtained by using the softmax function to the logit. 
As shown in Figure~\ref{fig:short-b}, the positive correlation between the target class probability and input transformation diversity is still evident.
Based on the above analysis, we can conclude that transformed adversarial examples with high logit outputs are more likely to exhibit adversarial transferability in targeted attacks.

\subsection{Rethinking Operations Used in AutoAugment}

To improve the logit of the target class, we propose a new method called AutoAugment Input Transformation (AAIT) attack. 
The inspiration comes from the concept of AutoAugment~\cite{AutoAugment}, which creates a search space of data augmentation policies and searches for the optimal combination of these policies.
Significant progress has been made in the field of data augmentation for standard model training, as demonstrated by related works such as AutoAugment~\cite{AutoAugment}, Fast AutoAugment~\cite{Fast}, Faster AutoAugment~\cite{Faster}, and PBA~\cite{PBA}. 
These approaches have achieved remarkable results using various data augmentation techniques during training.
Our framework draws inspiration from the Faster AutoAugment approach, which utilizes backpropagation to expedite the search process. 
Specifically, we rethink the operations in the search space and redesign the objective function to align with the targeted attack requirements.
By leveraging these modifications, our AAIT method aims to improve logit of the target class, thereby enhancing the effectiveness of targeted attacks.

Before formally introducing our method, some basis should be described.
In our framework, input images are transformed by a policy that consists of \(L\) different sub-policies.
Each sub-policy has \(K\) consecutive image processing operations.
In previous work~\cite{AutoAugment}, operations can be categorized into two main types: affine transformations and color enhancing operations. 
Therefore, we conduct experiments under three settings to explore whether both categories are important in improving adversarial transferability.
Specifically, we use affine transformations and color enhancing operations separately to search for the best policy and integrate them into the combination of DIM, TIM, and MI-FGSM (DTMI).

Table~\ref{affine} reports the targeted attack success rate using DTMI, DTMI-Affine, and DTMI-Color.
It is surprising that color enhancing operations even exhibit a lower success rate compared to DTMI.
On the contrary, affine transformations can improve transferability when integrated with DTMI and outperform with a large margin of nearly 50\%. 
Therefore, the operations used in our search space only include affine transformations, which are ShearX, ShearY, TranslateX, TranslateY, Rotate and Flip.

\begin{table}[htbp]
\caption{The targeted attack success rates (\%) of DTMI integrated with affine and color transformations. The adversarial examples are crafted on ResNet-50 model. The best result in each column is in \textbf{bold}.}
\label{affine}
\begin{center}
\begin{tabular}{l|cccc}
\toprule
Attack  &DN-121  &VGG-16  &Inc-v3 \\
\midrule
DTMI         &72.6  &62.2  &~~9.7  \\
DTMI-Affine  &\textbf{91.5}  &\textbf{86.6}  &\textbf{59.5} \\
DTMI-Color   &67.2  &56.0  &15.0 \\
\bottomrule
\end{tabular}
\end{center}
\end{table}


\renewcommand{\algorithmicrequire}{ \textbf{Input:}}
\renewcommand{\algorithmicensure}{ \textbf{Output:}} 

\begin{algorithm}[t]
\caption{The AAIT search algorithm}
\label{alg:search}
\begin{algorithmic}[1]
\REQUIRE  A classifier $f$ with loss function $J$, dataset $\mathcal{D}$ with target labels, distance function \(D(\cdot,\cdot)\)\\
\REQUIRE Distance between two densities $d$, classification loss coefficient $\eta$
\ENSURE A policy with probability parameters $p$ and magnitude parameters $\mu$\\ 
    \STATE Initialize a policy with $L$ sub-policies
    \WHILE{not converge}
      \STATE Sample a batch images from $\mathcal{D}$ and divide them into two parts $\mathcal{A},\mathcal{B}$
      \STATE Random select a sub-policy from policy to transform images $\mathcal{A}^{\prime} = S(\mathcal{A};\mu, p) $
      \STATE Calculate distance $d=D(\mathcal{A}^{\prime}, \mathcal{B})$
      \STATE Calculate classification loss $l=$\\$\mathbb{E}_{(X,y_t)\sim\mathcal{A}^{\prime}}J(f(X),y_t)+\mathbb{E}_{(X^{\prime},y^{\prime}_t)\sim\mathcal{B}}J(f(X^{\prime}),y_{t}^{\prime})$
      \STATE Update parameters $p, \mu$ to minimize $d + \eta l$ using stochastic gradient descent 
    \ENDWHILE
    \RETURN An optimal policy with probability parameters $p$ and magnitude parameters $\mu$
\end{algorithmic}
\end{algorithm}
\subsection{The New Goal for Targeted Attacks}

To search for the optimal policy for targeted attacks, we design two components as the goal of the search process.
First, the above analysis suggests maximizing the logit of the target class on transformed images, which implies a greater likelihood of generating highly transferable examples.
Then, we minimize the distributional distance between the original and transformed images.
In this way, the transformed images can preserve important semantic information as much as possible while approaching the target class.
The search process of our algorithm is provided in Algorithm~\ref{alg:search}.

After the search process, we get the optimal policy and denote it as the function $Policy(\cdot)$ to transform images.
Considering both the random selection of the sub-policy and the probability \(p\) of the operations, we generate a set of transformed images and obtain the average gradients on them in~\cref{at}.

\begin{equation} \label{at}
\overline{\boldsymbol{g}}_{t+1}=\frac{1}{m}\sum_{i=0}^{m-1}\nabla_{\boldsymbol{x}_{t}^{adv}}\mathcal{L}(\textit{Policy}(\boldsymbol{x}_{t}^{adv}),y_t)),
\end{equation}
where $m$ is the number of transformed images.
We will provide the complete AAIT attack algorithm in Appendix~\ref{A}.
By the automatic search with a powerful objective, AAIT can make the adversarial examples move toward the target classes during the iterations and further boost targeted transferability. 
\begin{table*}[t]
\renewcommand{\arraystretch}{1.2}
\caption{The targeted attack success rates (\%) against ten CNN-based models on the ImageNet-Compatible dataset. The best result in each column is in \textbf{bold}.}
\label{single}
\begin{center}
\resizebox{\textwidth}{!}{
\begin{tabular}{lccccccccccc}
\toprule[1.6pt]
\textbf{Source:DN-121}  &\multicolumn{10}{c}{Target model}  \\
\cmidrule(lr){2-11}
Attack &VGG-16 &RN-18 &RN-50 &DN-121 &Xcep &MB-v2 &EF-B0 &IR-v2 &Inc-v3 &Inc-v4 &Avg.\\
\midrule[0.8pt]
DTMI   &37.7&31.7&45.6&98.9&~~4.2&13.2&17.8&~~4.9&~~8.0&~~8.3&27.03\\
SI-DTMI &40.9&43.3&51.0&\textbf{99.1}&10.5&17.8&28.5&11.1&22.1&15.7&34.00\\
Admix-DTMI &46.8&46.6&55.7&98.7&11.3&21.0&31.4&14.0&24.6&19.2&36.93\\
ODI-DTMI 
&67.5&64.6&74.1&97.1&33.3&47.6&56.2&38.4&52.1&46.5&57.74\\
\rowcolor{gray!20}
    \textbf{AAIT-DTMI}   &\textbf{74.0}&\textbf{74.8}&\textbf{81.5}&97.7&\textbf{41.0}&\textbf{59.2}&\textbf{66.9}&\textbf{39.2}&\textbf{54.5}&\textbf{52.3}&\textbf{64.11}\\
\bottomrule[0.8pt]
\textbf{Source:Inc-v3}  &\multicolumn{10}{c}{Target model}  \\
\cmidrule(lr){2-11}
Attack &VGG-16 &RN-18 &RN-50 &DN-121 &Xcep &MB-v2 &EF-B0 &IR-v2 &Inc-v3 &Inc-v4 &Avg.\\
\midrule[0.8pt]
DTMI   &~~3.1&~~2.2&~~2.8&~~4.2&~~2.1&~~1.4&~~3.4&~~2.5&98.9&~~4.1&12.47\\
SI-DTMI    &~~4.0&~~5.6&~~5.6&11.7&~~5.6&~~3.9&~~7.3&~~8.1&\textbf{99.1}&10.9&16.18\\
Admix-DTMI &~~5.6&~~7.7&~~8.4&13.5&~~5.9&~~4.5&~~8.9&11.9&\textbf{99.1}&13.6&17.91\\
ODI-DTMI &16.9&15.7&19.8&34.5&23.1&15.4&24.8&26.4&95.8&32.3&30.46\\
\rowcolor{gray!20}
\textbf{AAIT-DTMI}   &\textbf{20.3}&\textbf{23.8}&\textbf{23.6}&\textbf{39.5}&\textbf{29.8}&\textbf{17.9}&\textbf{29.4}&\textbf{35.5}&98.6&\textbf{42.5}&\textbf{36.09}\\
\bottomrule[1.2pt]

\end{tabular}}
\end{center}
\end{table*}
\section{Experiments}
\subsection{Experimental Setup}
\textbf{Datasets and Models.}
To align with previous works~\cite{CFM}, we follow their settings. 
Specifically, the dataset is ImageNet-Compatible dataset, which is released for the NIPS 2017 adversarial attack challenge and consists of
1000 images and corresponding labels for targeted attacks. 
We attack ten normal trained models: VGG-16~\cite{VGG}, ResNet-18 (RN-18)~\cite{resnet}, ResNet-50 (RN-50)~\cite{resnet}, DenseNet-121 (DN-121)~\cite{densenet}, Xception (Xcep)~\cite{Xception}, MobileNet-v2 (MB-v2)~\cite{MobileNetV2}, EfficientNet-B0 (EF-B0)~\cite{EfficientNet}, Inception ResNet-v2 (IR-v2)~\cite{Inception-ResNet}, Inception-v3 (Inc-v3)~\cite{Inceptionv3}, and Inception-v4 (Inc-v4)~\cite{Inception-ResNet}. 
The architectures of these models are diverse.
Therefore, it is more challenging to attack.
Additionally, we add five Transformer-based models: Vision Transformer (ViT)~\cite{ViT}, LeViT~\cite{LeViT}, ConViT~\cite{ConViT}, Twins~\cite{Twins}, and Pooling-based Vision Transformer (PiT)~\cite{PiT}.
We also evaluate the proposed method on the CIFAR-10 dataset.
The images and target classes are provided by~\cite{CFM}.
Besides the normal trained models, we also use several ensemble models composed of three ResNet-20~\cite{resnet} networks (ens3-RN-20).
They are trained under three settings: standard training, ADP~\cite{ADP}, and DVERGE~\cite{DVERGE}.

\textbf{Baselines.}
We compare the targeted attack success rates with some advanced input transformation attacks and their various combinations, including DIM~\cite{DI}, MI-FGSM~\cite{MIM}, TIM~\cite{TIM}, SIM~\cite{NI}, Admix~\cite{Admix} and ODI~\cite{ODI}.
Particularly, we adopt the combination of DIM, TIM, and MI-FGSM (DTMI) as the baseline. 
Other methods, including our AAIT method, are integrated with DTMI to improve attack performance. 

\textbf{Attack setting.}
We follow the setting of previous work~\cite{CFM}.
Specifically, we set the \(l_\infty\)-norm perturbation boundary \(\epsilon=16/255\) and the step size \(\alpha=2/255\). 
We set the total iterations $T$ to 300. 
All the methods adopt the simple logit loss~\cite{Logit_loss} to optimize the adversarial examples.
We set the decay factor \(\mu=1.0\) for all the methods. 
For DIM
, we set the transformation probability \(p=0.7\), and the images are maximally resized to \(330 \times 330\).
For TIM
, we use the \(5 \times 5\) Gaussian kernels.
For SIM
and Admix, we follow the default settings and change the number of scale copies to 5 and mixed images to 3 for a fair comparison. 
The mixing weight for Admix is set to 0.2.
For ODI, we follow~\cite{ODI} for the detailed setup.
For our method, AAIT, a searched policy comprises 10 sub-policies, each with 2 operations. We train 20 epochs to search and set classification loss coefficient \(\eta\) to 0.3. When crafting adversarial examples, we average the gradient on 5 transformed images. More details will be provided in Appendix~\ref{setting}.



\begin{table*}[t]
\renewcommand{\arraystretch}{1.2}
\caption{The targeted attack success rates (\%) against five Transformer-based models on the ImageNet-Compatible dataset. The best result in each column is in \textbf{bold}.}
\label{vit}
\begin{center}
\resizebox{\textwidth}{!}{
\begin{tabular}{lcccccc|lcccccc}
\toprule[1.6pt]
\textbf{Source:DN-121}  &\multicolumn{5}{c}{Target model}  &\multicolumn{1}{c}{}&\textbf{Source:Inc-v3}  &\multicolumn{5}{c}{Target model}\\
\cmidrule(lr){2-6} \cmidrule(lr){9-13} 
Attack &ViT &LeViT &ConViT &Twins &PiT &Avg.&Attack&ViT &LeViT &ConViT &Twins &PiT &Avg.\\
\midrule[0.6pt]
DTMI   &~~0.2&~~2.5&0.3&~~0.9&~~1.0&0.98&DTMI&0.1&~~0.3&0.0&0.1&0.0&0.10\\
SI-DTMI &~~1.6&~~6.2&0.7&~~2.2&~~4.1&2.96&SI-DTMI&1.0&~~2.5&0.1&0.5&0.7&0.96\\
Admix-DTMI  &~~1.9&~~8.6&1.2&~~2.8&~~5.3&3.96&Admix-DTMI  &0.5&~~3.3&0.5&1.1&1.6&1.40\\
ODI-DTMI    &~~3.1&28.9&8.5&15.1&22.0&15.52&ODI-DTMI&1.0&13.1&2.1&4.7&8.3&5.84\\
\rowcolor{gray!20} \textbf{AAIT-DTMI}   &\textbf{11.7}&\textbf{35.6}&\textbf{9.0}&\textbf{19.6}&\textbf{25.7}&\textbf{20.32}&\textbf{AAIT-DTMI}&\textbf{4.2}&\textbf{18.0}&\textbf{2.4}&\textbf{5.8}&\textbf{9.7}&\textbf{8.02}\\
\bottomrule[1.6pt]
\end{tabular}}
\end{center}
\end{table*}
\subsection{Performance Comparison}

\textbf{Attack on CNN-based models.} Table~\ref{single} shows the targeted attack success rates against ten black-box models with DN-121 and Inc-v3 as the source models.
AAIT outperforms all baselines with a clear margin in all the source models.
Specifically, the average attack success rate increases by 6.37\% over the second-best method when the source model is DN-121.
Moreover, when the substitution model is Inc-v3, the average attack success rate of AAIT also achieves 36.09\%, gaining an improvement of 5.63\%.
Compared with ODI, AAIT has better adversarial transferability while maintaining a high white-box attack success rate.

\textbf{Attack on Transformer-based models.}
Table~\ref{vit} reports the attack success rates against five Transformer-based models. 
DN-121 and Inc-v3 are also chosen to be the source models.
As reported in Table~\ref{vit}, all the baseline methods exhibit low attack success rates when adversarial examples transfer to the ViT-based model.
On the contrary, our proposed AAIT attack boosts the attack success rate from 3.1\% to 11.7\% (3$\times$) and from 1.0\% to 4.1\% (4$\times$).
The average targeted attack success rate also increases.
Specifically, the average targeted attack success rate approaches 20.32\% and 8.02\%, gaining an improvement of 4.80\% and 2.18\%, respectively.
It indicates that our AAIT method achieves higher attack success rates compared to other target attack methods, even when the model architecture is different.

\begin{table*}[t]
\renewcommand{\arraystretch}{1.2}
\caption{The targeted attack success rates (\%) against normal trained and ensemble models on the CIFAR-10 datasst. The best result in each column is in \textbf{bold}.}
\label{cifar}
\begin{center}
\begin{tabular}{lcccccccccc}
\toprule[1.6pt]
\textbf{Source:RN-50} &\multicolumn{9}{c}{Target model}  \\
\cmidrule(lr){2-10}
\multirow{2}{*}{Attack} &\multirow{2}{*}{VGG-16} &\multirow{2}{*}{RN-18} &\multirow{2}{*}{RN-50} &\multirow{2}{*}{MB-v2} &\multirow{2}{*}{Inc-v3} &\multirow{2}{*}{DN-121} &\multicolumn{3}{c}{ens3-RN-20} &\multirow{2}{*}{Avg.}\\
&&&&&&&Baseline &ADP &DVERGE \\
\midrule[0.8pt]
DTMI   &65.8&71.8&\textbf{100.0}&62.6&74.1&83.3&77.3&57.0&13.2&67.23\\
SI-DTMI    &72.6&75.9&\textbf{100.0}&74.8&75.5&85.6&79.6&62.5&21.3&71.98\\
Admix-DTMI  &81.7&82.8&99.9&82.5&82.3&90.7&85.6&68.0&25.6&77.68\\
ODI-DTMI    &87.3&88.9&97.5&82.6&88.9&91.4&88.6&79.9&\textbf{47.8}&83.66\\
\rowcolor{gray!20}
\textbf{AAIT-DTMI}   &\textbf{93.1}&\textbf{94.3}&\textbf{100.0}&\textbf{90.1}&\textbf{92.7}&\textbf{96.4}&\textbf{94.5}&\textbf{86.3}&42.7&\textbf{87.79}\\
\bottomrule[1.6pt]
\end{tabular}
\end{center}
\end{table*}

\textbf{Attack normal trained and ensemble models.}
We also evaluate the adversarial transferability
on the CIFAR-10 dataset.
Table~\ref{cifar} reports the targeted attack success rates against six non-robust models and three different ResNet-20 ensemble models with adversaries crafted on the ResNet-50 model.
Compared to other methods, AAIT not only exhibits a higher attack success rate, but also behaves well under the white-box setting. 
Specifically, AAIT increases the average attack success rate by 4.13\% over ODI and achieves a 100\% attack success rate while transferring to RN-50.

\begin{figure*}[htbp]
  \centering
   \includegraphics[width=\linewidth]{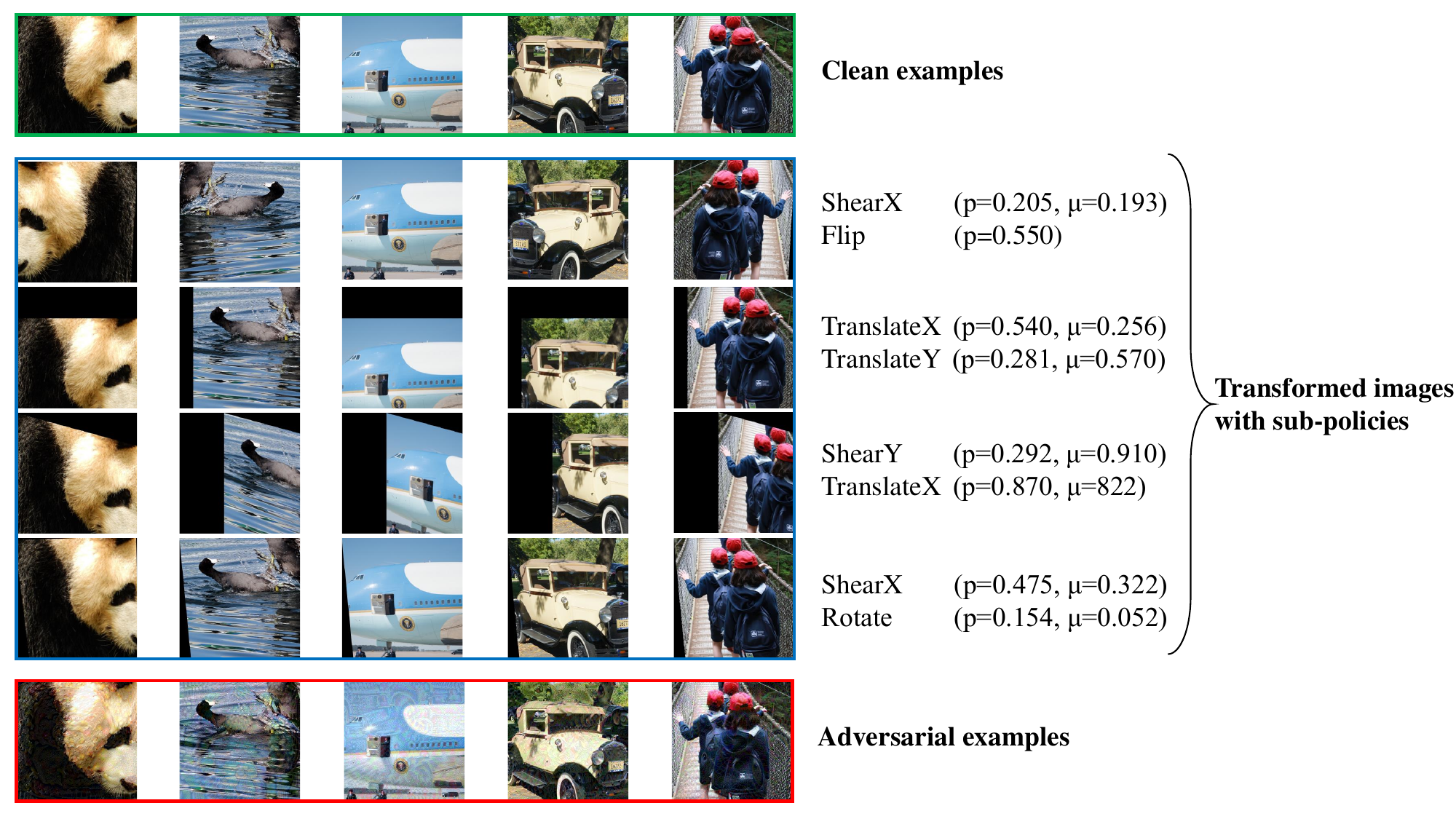}
   \caption{The clean examples and transformed images with four selected sub-policies. AAIT creates diverse transformed images with randomly selected sub-policies on the right side. The transformed images are different even with the same sub-policy because of the probability $p$. The adversarial examples are also presented.
   }
   \label{fig:aug}
\end{figure*}

\textbf{Searched policy on Imagenet-Compatible dataset.} In Figure~\ref{fig:aug}, we show four sub-policies and corresponding transformed images. 
The sub-policies learn different operations and parameters, which makes transformed images more diverse.
And these images also preserve important semantic information, easily recognized by humans.

\subsection{Ablation Study}
We conduct experiments to investigate the impact of transformed image number $m$ and classification loss coefficient $\eta$.
For these ablation experiments, we select five CNN-based and Transformer-based models.
\begin{table*}[t]
\renewcommand{\arraystretch}{1.2}
\caption{Comparison of targeted attack success rates (\%) with Logit and CE losses in search algorithm. The best result in each column is in \textbf{bold}.}
\label{loss}
\begin{center}
\resizebox{\textwidth}{!}{
\begin{tabular}{cccccccccccccc}
\toprule[1.6pt]
\multirow{2}{*}{Source model}&\multicolumn{2}{c}{Loss}  &\multicolumn{10}{c}{Target model}  \\
\cmidrule(lr){2-3}\cmidrule(lr){4-13}
&Logit &CE &VGG-16 &RN-50 &Xcep &MB-v2 &EF-B0 &Inc-v4 &ViT &LeViT &ConViT &PiT &Avg.\\
\midrule[0.6pt]
\multirow{2}{*}{DN-121}
&&\checkmark  &\textbf{76.5}&\textbf{83.4}&38.7&58.2&64.5&49.6&9.8&32.6&5.9&19.8&43.90\\
&\checkmark&  &74.0&81.5&\textbf{41.0}&\textbf{59.2}&\textbf{66.9}&\textbf{52.3}&\textbf{11.7}&\textbf{35.6}&\textbf{9.0}&\textbf{25.7}&\textbf{45.69}\\
\midrule[0.6pt]
\multirow{2}{*}{Inc-v3}
&&\checkmark  &\textbf{22.3}&23.5&24.1&15.5&27.8&38.5&2.6&14.6&\textbf{2.7}&7.9&17.95\\
&\checkmark&
&20.3&\textbf{23.6}&\textbf{29.8}&\textbf{17.9}&\textbf{29.4}&\textbf{42.5}&\textbf{4.2}&\textbf{18.0}&2.4&\textbf{9.7}&\textbf{19.78}\\
\bottomrule[1.6pt]

\end{tabular}}
\end{center}
\end{table*}

\begin{figure}[t]
  \centering
  \begin{subfigure}{0.48\linewidth}
    \includegraphics[width=1\linewidth]{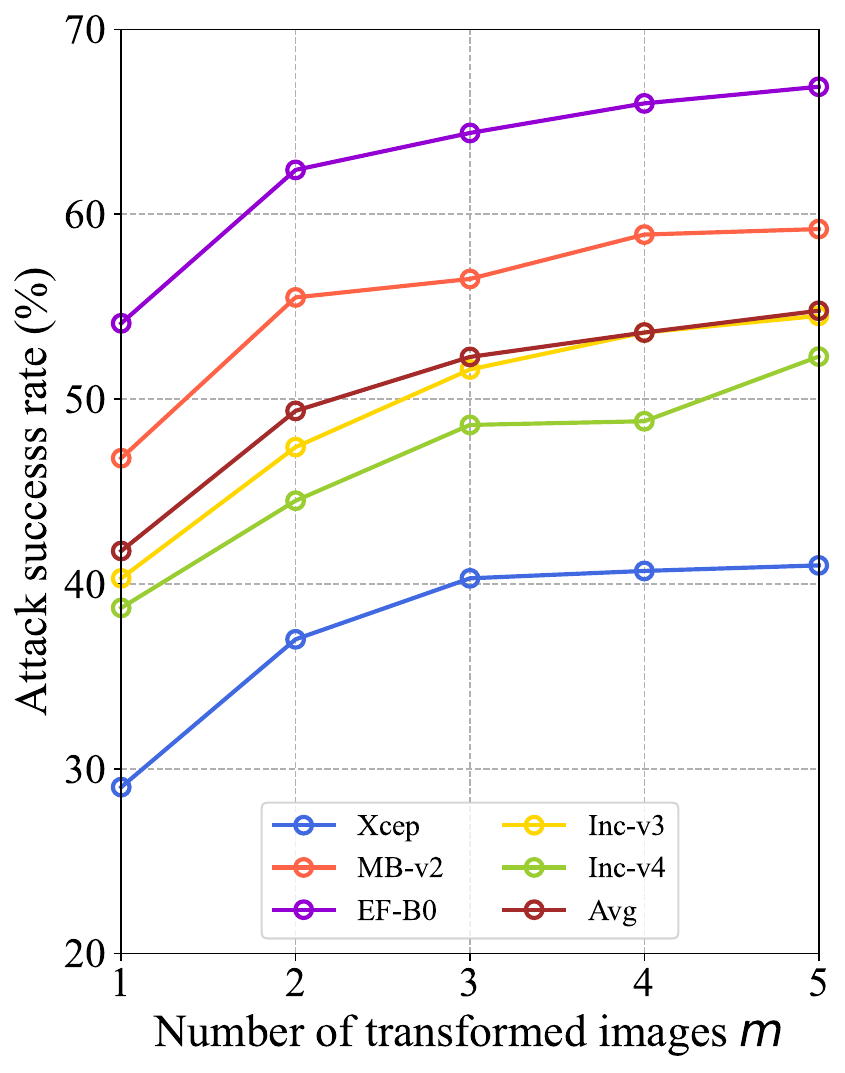}
    \caption{CNN-based models}
    \label{fig:m-a}
  \end{subfigure}
  \hfill
  \begin{subfigure}{0.48\linewidth}
    \includegraphics[width=1\linewidth]{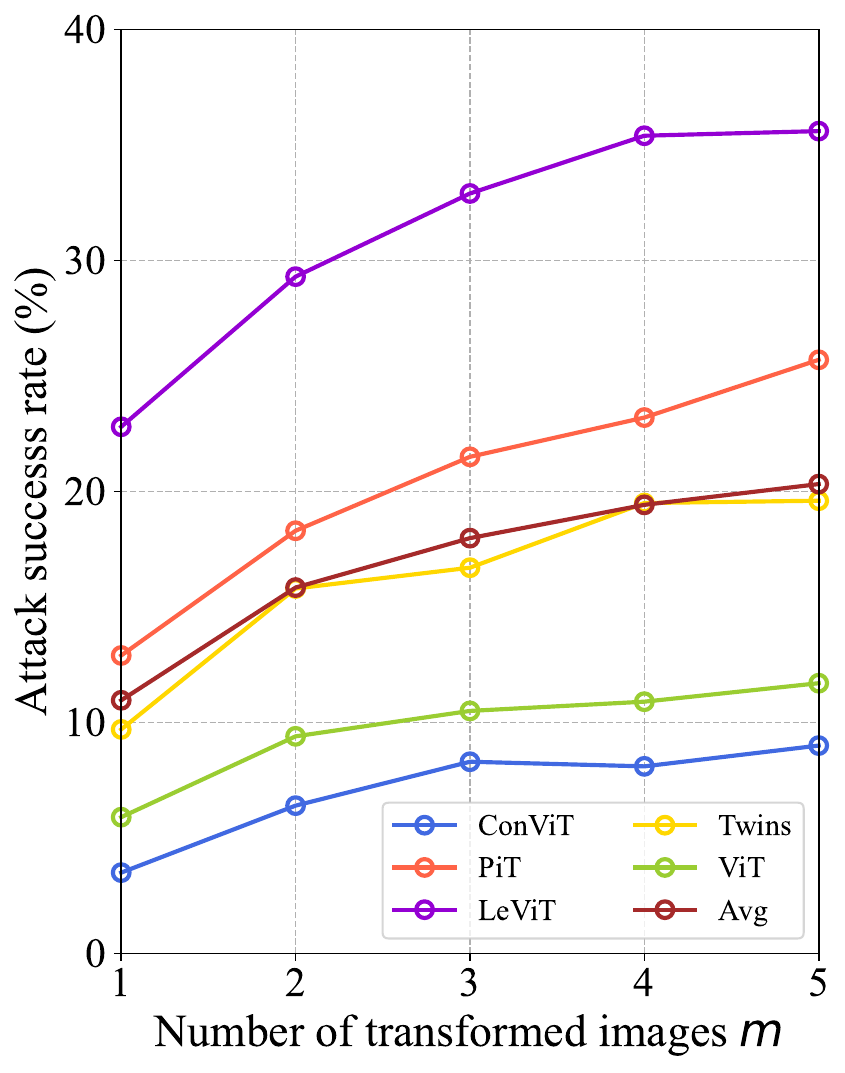}
    \caption{Transformer-based classifiers}
    \label{fig:m-b}
  \end{subfigure}
  \caption{The targeted attack success rates on both CNN-based and Transformer-based models with adversaries crafted on DN-121 model for various transformed images number $m$.}
  \label{fig:m}
\end{figure}

\textbf{Transformed image number $m$.} 
As illustrated in Figure~\ref{fig:m}, we show the targeted attack success rate of AAIT on both CNN-based and Transformer-based models with adversaries crafted on the DN-121 model, where $\eta$ is fixed to 0.3. 
When $m < 3$, it is evident that the attack success rate of adversarial examples on all models significantly improves as the value of $m$ increases.
When $m > 3$, there is a slight tendency for the attack success rate to increase.
Specifically, when the value of $m$ is increased from 4 to 5, only two models (Inc-v4 and PiT) demonstrate an increase in the attack success rate.
Thus, we set $m=5$ to balance increased time overhead and attack performance.

\begin{figure}[t]
  \centering
  \begin{subfigure}{0.48\linewidth}
    \includegraphics[width=1\linewidth]{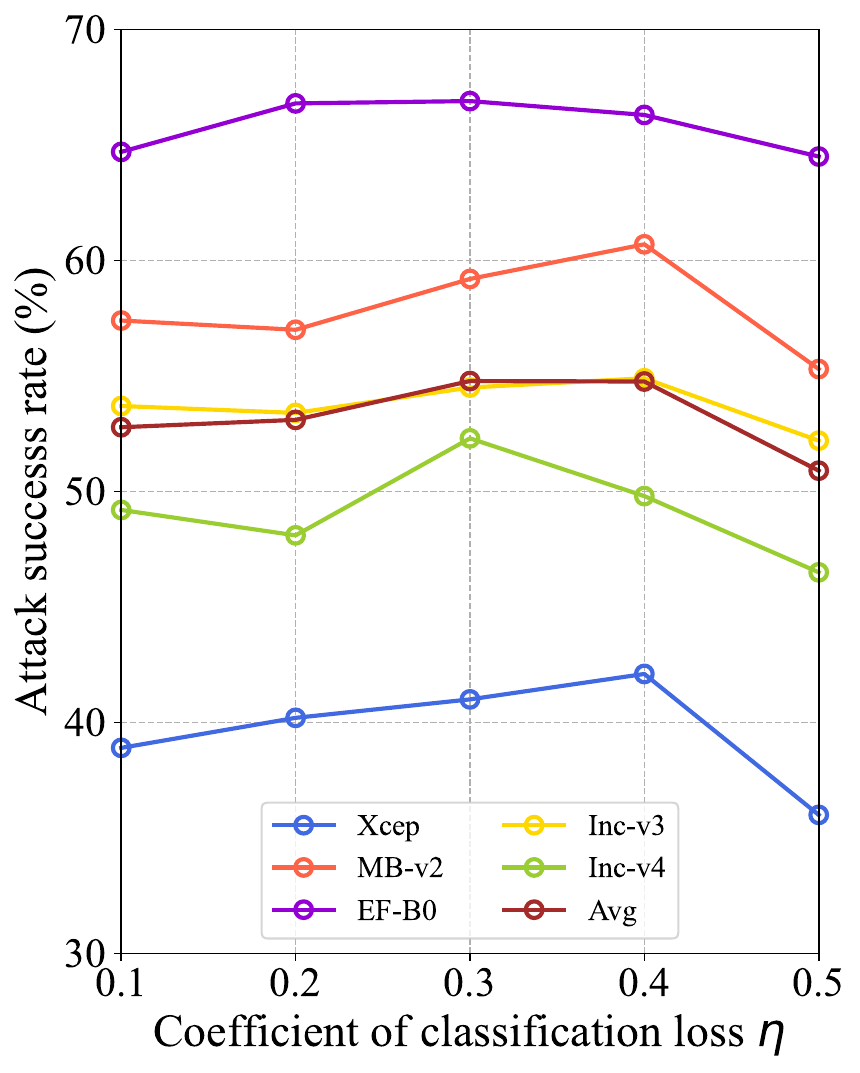}
    \caption{CNN-based models}
    \label{fig:epsilon-a}
  \end{subfigure}
  \hfill
  \begin{subfigure}{0.48\linewidth}
    \includegraphics[width=1\linewidth]{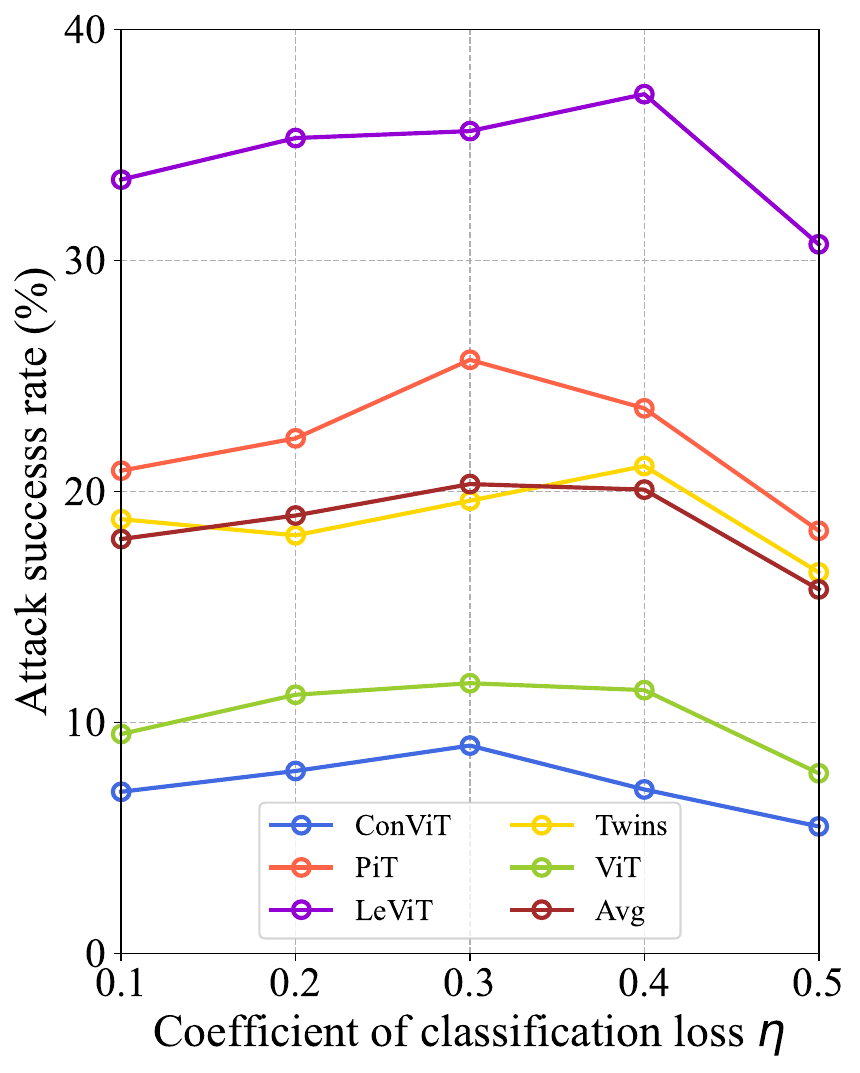}
    \caption{Transformer-based classifiers}
    \label{fig:epsilon-b}
  \end{subfigure}
  \caption{The targeted attack success rates on both CNN-based and Transformer-based models with adversaries crafted on DN-121 model for various classification loss coefficient $\eta$.}
  \label{fig:eta}
\end{figure}

\textbf{Classification loss coefficient $\eta$.} 
As shown in Figure~\ref{fig:eta}, we illustrate the targeted attack success rate of AAIT with adversaries crafted on the DN-121 model, where $m$ is fixed to 5. 
As the value of $\eta$ increases, the attack success rate of adversarial examples also increases. 
For certain models, the attack success rate reaches the peak when $\eta=0.3$, while for others, the peak is achieved at $\eta=0.4$. 
This implies that the optimal choice of $\eta$ varies depending on the specific model under consideration.
To provide a comprehensive analysis, we also report the average attack success rate.
It is observed that when $\eta=0.3$, both CNN-based and Transformer-based models exhibit a slightly higher success rate compared to that when $\eta=0.4$. 
We eventually set $\eta=0.3$ as it generally offers improved performance.

\textbf{Search algorithm loss.} 
In Table~\ref{loss}, we also evaluate how the attack success rate varies using two classification losses in the search algorithm.
Here $m$ is fixed to 5, and $\eta$ is set to 0.3.
One can observe that, when using Logit loss, most target models exhibit a higher attack success rate than CE loss.
The average success rate also shows the advantage of Logit loss, which is eventually used in the search algorithm.
\section{Conclusion}
In this work, we propose a novel targeted adversarial attack method based on input transformation techniques, called AutoAugment Input Transformation (AAIT). 
Compared with other input transformation methods, our proposed AAIT method leverages a search process to identify the optimal transformation policy. 
This is achieved by maximizing the logit of transformed examples while minimizing the distribution distance between the clean and transformed examples.
Extensive experiments conducted on CIFAR-10 and ImageNet-Compatible datasets demonstrate that AAIT outperforms other transfer-based targeted attacks.

Our results showcase the superior performance of AAIT in generating highly transferable adversarial examples.
However, compared to non-targeted attacks, the transferability in targeted attacks is still far from the saturation. 
We hope our work could inspire more research in the future.
{
    \small
    \bibliographystyle{ieeenat_fullname}
    \bibliography{main}
}

\clearpage
\section*{Appendix}
\appendix
\section{AAIT Operations and Descriptions} 
Here, we provide both affine and color operations and their descriptions in Table~\ref{operations}.
Note that the upper part is affine transformations in our search space and the lower part is color transformations.
\begin{table}[htbp]
\renewcommand{\arraystretch}{1.2}
\caption{The affine and color transformations used in AAIT and corresponding descriptions.}
\label{operations}
\begin{center}
\begin{tabular}{l|l}
\toprule
Operation  &Description   \\
\midrule
\multirow{2}{*}{ShearX/Y}     &Shear the image along the \\&horizontal (vertical) axis.     \\
\multirow{2}{*}{TranslateX/Y}  &Translate the image in the \\&horizontal (vertical) direction.     \\
Rotate        &Rotate the image magnitude degrees.     \\
Flip          &Flip an image along the horizontal axis.     \\ 
\hline
Solarize &Invert all pixels above a threshold value.\\
Posterize &Reduce the number of bits for each pixel.\\
Invert &Invert the pixels of the image.\\
Contrast &Control the contrast of the image.\\
Color &Adjust the color balance of the image.\\
Brightness &Adjust the brightness of the image.\\
Sharpness &Adjust the sharpness of the image.\\
\multirow{3}{*}{AutoContrast} &Maximize the the image contrast, by \\&making the darkest pixel black \\&and lightest pixel white.\\
Equalize &Equalize the image histogram.\\
\bottomrule
\end{tabular}
\end{center}
\end{table}

\section{Algorithm}
\label{A}
It is worth noting that our proposed  AAIT method can be integrated with existing gradient-based attacks and other input transformation methods.
Specifically, AAIT can combined with DIM~\cite{DI}, TIM~\cite{TIM} and MI-FGSM~\cite{MIM} to AAIT-DI-TI-MI-FGSM.
The pseudo-codes of AAIT-DI-TI-MI-FGSM attack are described in Algorithm~\ref{alg:AAIT}.

\renewcommand{\algorithmicrequire}{ \textbf{Input:}}
\renewcommand{\algorithmicensure}{ \textbf{Output:}} 

\begin{algorithm}[htbp]
\caption{The AAIT-DI-TI-MI-FGSM attack algorithm}
\label{alg:AAIT}
\begin{algorithmic}[1]
\REQUIRE  
    A clean example $\boldsymbol{x}$; a target label $y_t$; a classifier $f$.\\
\REQUIRE Adversary loss funcion $\mathcal{L}(\cdot,\cdot)$;  perturbation boundary $\epsilon$; step size $\alpha$; maximum iterations $T$; decay factor $\mu$; Gaussian kernel $\boldsymbol{W}$.
\REQUIRE An optimal policy function $Policy(\cdot)$ searched by Algorithm~\ref{alg:search}
\ENSURE An adversarial example $\boldsymbol{x}^{adv}$~~\\ 
    \STATE $\boldsymbol{g}_0=0;\boldsymbol{x}_0^{adv}=\boldsymbol{x}$
    \FOR{$t=0 \rightarrow T-1$}
    \STATE First apply DIM to transform images:\\
           $\boldsymbol{x}^{adv}_{t} = DI(\boldsymbol{x}^{adv}_{t})$
    \STATE Calculate the gradient $\overline{\boldsymbol{g}}_{t+1}$ by \cref{at}\\
    \STATE Update the enhanced momentum by applying MI:\\
           \begin{equation}
               \boldsymbol{\Tilde{g}}_{t+1} = \mu \cdot \boldsymbol{g}_{t} + \frac{\overline{\boldsymbol{g}}_{t+1}}{\left\|\overline{\boldsymbol{g}}_{t+1}\right\|_1}
           \end{equation}
    \STATE Apply TIM by convolving the gradient:\\
    $\boldsymbol{g}_{t+1} = \boldsymbol{W} \ast \boldsymbol{\Tilde{g}}_{t+1}$
    \STATE $\boldsymbol{x}^{adv}_{t+1} = \boldsymbol{x}^{adv}_{t} - \alpha \cdot \text{sign}(\boldsymbol{g}_{t+1})$ 
    \STATE $\boldsymbol{x}^{adv}_{t+1} = Clip^{\epsilon}_{\boldsymbol{x}}(\boldsymbol{x}^{adv}_{t+1})$
    \ENDFOR
    \STATE $\boldsymbol{x}^{adv}=\boldsymbol{x}_T^{adv}$
    \RETURN $\boldsymbol{x}^{adv}$
\end{algorithmic}
\end{algorithm}

\section{Details of Attack Settings}
\label{setting}

Here, we provide the details for our proposed AAIT method.
The searched optimal policy is composed of 10 sub-policies and each of them consists of 2 operations.
In the search process, we set temperature parameters to 0.05, which is used for differentiable search in~\cite{Faster}.
We use Adam optimizer with a learning rate of $1.0^{-3}$, coefficients for running averages (betas) of (0, 0.999), and train for 20 epochs.
The classification loss coefficient is set to 0.3 and the chunk size is set to 8 for all datasets.
The distance function is Wasserstein distance using Wasserstein GAN with gradient penalty~\cite{Faster}.
For generating adversarial examples, we use searched optimal policy to transform 5 images, keeping balance between effectiveness and efficiency.
Then, we obtain the average gradient from transformed images to update adversarial perturbations.

\section{Visualization of Adversarial Examples}
\begin{figure*}[htbp]
  \centering
   \includegraphics[width=0.75\linewidth]{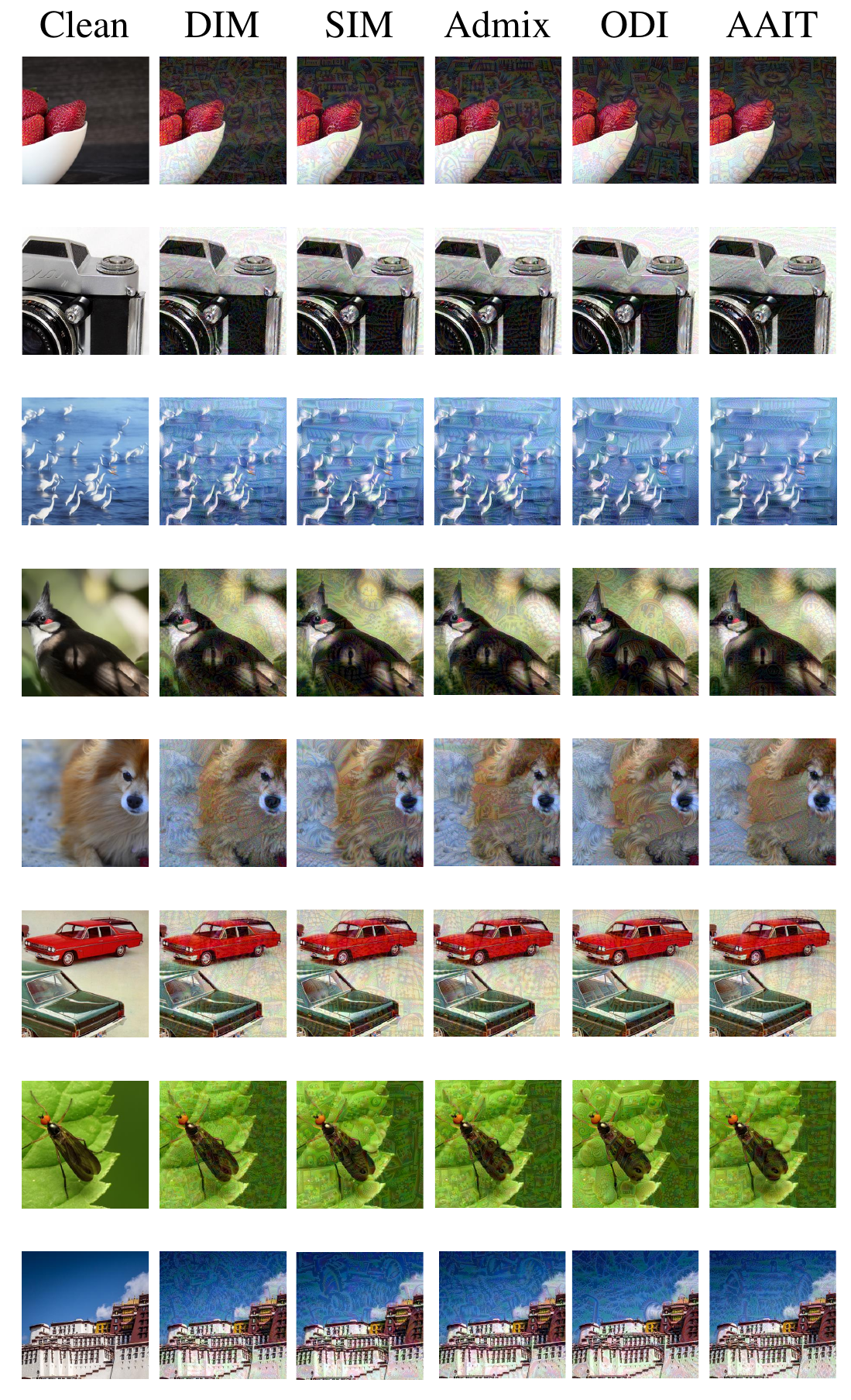}
   \caption{Visualization of generated adversarial examples.
   }
   \label{fig:appendix}
\end{figure*}

In Figure~\ref{fig:appendix}, we visualize 8 randomly selected clean images and their corresponding adversarial examples with different input transformation methods. The source model is DenseNet-121. Compared with other advanced adversarial attack, our generated adversarial perturbations are also human imperceptible.

\label{Appendix}


\end{document}